\documentclass{article}
\usepackage{spconf,amsmath,graphicx}
\usepackage{amsmath}
\usepackage{amsfonts}
\usepackage{multirow}
\usepackage{xcolor}
\usepackage{xspace}
\usepackage{url}
\usepackage[normalem]{ulem}

\newcommand{\IndNet}{Ind-Net\xspace}
\newcommand{\PredNet}{Pred-Net\xspace}
\newcommand{\MultiNet}{MultiFrame-Net\xspace}

\newcommand\blfootnote[1]{%
  \begingroup
  \renewcommand\thefootnote{}\footnote{#1}%
  \addtocounter{footnote}{-1}%
  \endgroup
}

\title{Learning Double-Compression Video Fingerprints\\ Left from Social-Media Platforms}
\name{Irene Amerini$^1$\qquad  Aris Anagnostopoulos$^{1,*}$\qquad Luca Maiano$^{1,2}$\qquad Lorenzo Ricciardi Celsi$^{1,2}$}
\address{Sapienza University of Rome$^1$\hspace{3cm}
ELIS Innovation Hub$^2$}
\begin{document}
%
\maketitle
\begin{abstract}
Social media and messaging apps have become major communication
platforms. Multimedia contents promote improved user engagement and have
thus become a very important communication tool. However, fake news and
manipulated content can easily go viral, so, being able to verify the source
of videos and images as well as to distinguish between native and
downloaded content becomes essential. Most of the work performed so far on
social media provenance has concentrated on images; in this paper, we propose
a CNN architecture that analyzes video content to trace videos
back to their social network of origin.  The experiments demonstrate
that stating platform provenance is possible for videos as well as
images with very good accuracy.
\end{abstract}
\section{Introduction}
\label{sec:intro}
\blfootnote{$\!\!\,^\ast$ Supported by the ERC Advanced Grant 788893 AMDROMA, the EC H2020 RIA project ``SoBigData++'' (871042), and the MIUR PRIN project ALGADIMAR.}

In recent years multimedia content has become one of the predominant
ways for exchanging
information. 
Every day people watch over a billion hours of video on
YouTube~\cite{youtube} and share more
than a billion stories on
Facebook~\cite{facebook}. The
expressiveness of visual content makes multimedia a powerful means of
communication. Therefore, it becomes increasingly important to be
able to verify the source of this information.

When uploaded and shared across social networks and messaging apps,
multimedia content undergoes a processing step in which the platforms
perform a set of operations on the input. Indeed, to optimize transfer
bandwidth as well as display quality, most platforms apply specific
compression and resizing methods. These methods, which tend to be
unpublished, differ among the different social platforms~\cite{giudiceACEIBSD2017}.
All these operations inevitably leave some
traces on the media content itself \cite{wangEDFVDDMC2006, stammTFAFMCV2012, longACBCNNFDDSVS2017}. The social media identification problem has been widely studied
for image files with promising results \cite{giudiceACEIBSD2017,
caldelliIOCBSNP2017, ameriniTIBSNOCBA2017}, employing machine learning
classifiers.  Recently, Quan et al.~\cite{quanPAIP2019} showed that by using
convolutional methods it is possible to recognize
Instagram filters and attenuate the sensor pattern noise
signal in images. Amerini et al.~\cite{ameriniSNITICWC2019} introduced a CNN
for learning distinctive features among social networks from the
histogram of the discrete cosine transform (DCT) coefficients and the
noise residual of the images.
Phan et al.~\cite{phanTMISSN2019} proposed a method to track multiple image sharing
on social networks by using a
CNN architecture able to learn a combination of DCT and metadata
features. Nevertheless, the identification of the traces left by
social networks and messaging apps on video contents remains an open
problem.
Recently, Iuliani et al.~\cite{iuluaniAVFFFUAMLF2019} presented an approach that relies 
on the analysis of the container structure of a video
through the use of unsupervised algorithms
 to perform source-camera identification
for shared media with high performance; their method is
strictly dependent on the file structure, whereas in our work we
are interested in approaches that are based on the content of a video,
independently of the file type. Kiegaing and Dirik~\cite{kiegaingPBSDAYV2019}
showed that
fingerprinting the I-frames of a flat content native video can be
used to accurately identify the source of YouTube videos.
Moreover, although the research community has treated video and image forensics
as separate problems, a recent work from
Iuliani et al. \cite{iulianiHRBVSI2019}, demonstrates that it is possible to
identify the source of a digital video by exploiting a reference
sensor pattern noise generated from still images taken by the same
device, suggesting that it could be possible to link social media
profiles containing images and videos captured by the same sensor.

In this work, we propose a multistream neural network architecture that
can capture the double compression traces left by social networks
and messaging apps on videos. 
According to our knowledge, this is the first work that investigates
whether it is possible to recognize videos from different social
networks by analyzing the traces of compression left by these sites when
loading content. The possibility of reconstructing information on the
sharing history of a certain object is highly valuable in media
forensics. In fact, it could help in monitoring the visual information
flow by tracing back the initial uploads, thus aiding source
identification by narrowing down the search. This could be helpful in
different applications such as, for example, cyberbullying, where we want
to be able to investigate who and where this individual has shared a certain content.
Similarly, this tool could be helpful to trace the sharing of videos of
military propaganda or other criminal activity back to the source, as
well as for fact checking and countering fake news. 

The problem of classifying
photos and videos from social networks has been typically treated
separately. To overcome this limitation, here we investigate
the possibility to test the robustness of our implementation with
respect to images once the network is trained on videos. The rest of
the paper is organized as follows: Section \ref{sec:method} describes
our approach. Section
\ref{sec:experimental-evaluation} discusses different experimental
results. Finally, Section \ref{sec:conclusions} draws the conclusions of
our work.

\section{Proposed Method}
\label{sec:method}

In video coding, a video is represented as a sequence of
\emph{groups of pictures} (GOP)s, each of which begins with an
\emph{I-frame}. I-frames are not predicted from any other frame and are
independently encoded using a process similar to JPEG compression. Apart
from the I-frames, the rest of each GOP consists of \emph{P-frames} and
\emph{B-frames}. These frames are predictively encoded using motion estimation
and compensation. Thus, these frames are derived from segments of an
anchor I-frame and represent lower quality frames. 
In this section we describe the proposed architecture (see
Figure~\ref{fig:architecture}) composed by a two-stream network, inspired
by the work by Nam et al.~\cite{namTSNFDDCHV2019}. 
However, the application of this particular network to the problem that we
study is novel and it requires some important modifications to the
method in \cite{namTSNFDDCHV2019}. First, we modified the third
convolutional block of the \IndNet removing a stack of Convolutional,
Batch Normalization, and ReLU operations and we added one more convolutional
block (Block 6) at the end of the CNN. This deeper configuration helps
the network to capture more subtle details in the input. Next, we
modified the \PredNet by doubling the number of operations in each block
and increased the number of output channels of each block in order to
learn a richer representation. Finally, we changed the dimensionality of
the flattened feature maps from 128 to 256 for the P-frames stream and
from 16,384 to 4,096 for the IF-stream. This helps to limit the importance
of I-frames over the P-frames.
We choose not to include B-frames in our analysis because of the
lower quality of these kind of frames. Finally, we introduce a
two-stream network (\MultiNet), which learns the inter-modal
relationships between features extracted from both types of frames. In
the rest of this section, we use the notation $W \times H$ to denote the resolution of a video $v$. Each video
can also be represented by $N$ frames denoted as
$f_{0}, \dots, f_{N-1}$, where
$f_{j} \in \mathbb{Z}^{3 \times W \times H}$.
Moreover, we use the notation $f_{Ii}^{(v)}$ and $f^{(v)}_{Pi}$ to
denote the $i$th I-frame or P-frame, respectively, of a video~$v$.

\subsection{\IndNet}
\label{sec:Ind-Net}
In this section we propose a network that analyzes the I-frames of a
video. The network is depicted in the bottom part of Figure
\ref{fig:architecture}. 
We designed a network that consists of six convolutional blocks that
act as a feature extractor and a fully connected network that takes the
input feature vector and produces an output classification. The first tree 
convolutional blocks made of (1) two consecutive stacks of
convolution (Conv2D), batch normalization (BatchNorm), and ReLU
operations, and (2) a final max pooling (MaxPool) layer. The last three
convolutional blocks are organized in three consecutive stacks of
(1) Conv2D, BatchNorm, and ReLU operations, and (2) a final MaxPool layer. Apart
from the first convolutional layer, which has a $5 \times 5$ kernel,
all other convolutional layers have a $3 \times 3$ kernel. The feature
extracted by the last MaxPool layer becomes eventually flattened and passed
through two stacks made by a 512-dimensional fully connected layer and
a ReLU, and a final 512-dimensional fully connected layer followed by a
softmax one. The network outputs a $\lvert C\rvert$-dimensional vector,
where $\lvert C\rvert$ is the number of output classes. 

Before being fed into the network, the decompressed I-frames are
initially transformed through a preprocessing module. To
highlight the traces left by double compression, we employ the high-pass
filter introduced by Fridrich and Kodovsky \cite[operator
S5a]{fridrichRMSDI2012}, and used in~\cite{namTSNFDDCHV2019}
and apply it to the Y-channel of the input after RGB-to-YUV
conversion. Therefore, we denote as $X_{Ii} = \{f^{(v)}_{Ii}\} \in
\mathbb{Z}^{3 \times W \times H}$ the input $i$th frame of video
\emph{v} and compute $X'_{Ii} = \{ \mathit{HPF}(Y(f^{(v)}_{Ii})) \} \in
\mathbb{Z}^{W \times H}$ to obtain the preprocessed input of the
network, where $\mathit{HPF}(\cdot)$ indicates the high pass filter and
$Y(\cdot)$ indicates the Y-channel of the input frame. Because we assume
that each video could come from a single social media platform, we train
the model using a cross-entropy loss function, thus training the model
to output a probability over the $\lvert C\rvert$ classes for each video.

\subsection{\PredNet}
\label{sec:pf-net}

\begin{figure*}[ht!]
    \centering
    \includegraphics[width=1.0\linewidth]{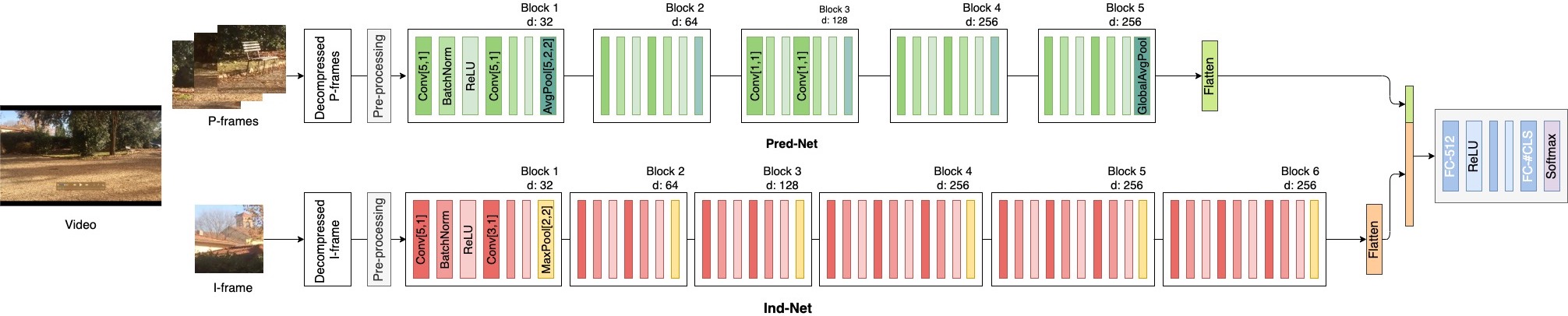}
    \caption{The proposed two-stream network (\MultiNet)
      architecture. The network is constructed by concatenating the
      feature maps of the \IndNet and the \PredNet. The I-frame and
      P-frame streams are trained separately. Next, we concatenate the
      flattened output of the two-streams and train a fully connected
      classifier.}
    \label{fig:architecture}
\end{figure*}

Now we present \PredNet, a new network architecture 
that analyzes the P-frames of a video to detect double
compression fingerprints. The network (depicted in the top of
Figure~\ref{fig:architecture}) is made of five convolutional blocks
and a fully connected network. All the convolutional blocks consist of
two stacks of (1) Conv2D, BatchNorm, and ReLU operations, and (2) a final average
pooling (AvgPool) layer. The AvgPool and GlobalAvgPool levels help to
preserve the statistical properties of feature maps that could otherwise
be distorted with the MaxPool. All the Conv2D layers in the first two
blocks have a $5 \times 5$ kernel, and the last three blocks have a
a $3 \times 3$ kernel. Finally, the feature maps extracted from the last
convolutional block are flattened and passed through a 256-dimensional
fully connected layer that outputs a $|C|$-dimensional vector and a
softmax operation that calculates the output prediction.

Similarly to the \IndNet, we add a preprocessing step to the input
frames in which a high-frequency--component extraction operation is
applied to eliminate the influence of diverse video contents.
Further, because the P-frames represent predicted low-quality frames, we
compensate for the loss of information by stacking consecutive frames.
In fact, given a stack of three consecutive P-frames denoted as
$X_{Pi} = \{ f^{(v)}_{Pi-1}, f^{(v)}_{Pi}, f^{(v)}_{Pi+1} \} \in \mathbb{Z}^{3 \times 3 \times W \times H}$,
we compute
$X'_{Pi} = \{ Y(f) - G(f) | f \in X_{Pi} \} \in \mathbb{R}^{3 \times W \times H}$,
where the function $G(\cdot)$ denotes a Gaussian filter. Like the
\IndNet, the network is trained with a cross-entropy loss function.

\subsection{\MultiNet}
\label{sec:tf-net}

Multistream architectures have been successfully applied by multimedia
forensics researchers for both forgery detection and source
identification tasks \cite{ameriniLJDCTMDCNN2017, verdeVCFBCNN2018,
mazaheriASCAFLIM2019, ameriniSNITICWC2019}. Therefore, we combine the
feature maps of both \IndNet and \PredNet to feed the fully connected
classifier with inter-modal relationships between different types of
frames. As shown in Figure~\ref{fig:architecture}, we concatenate the
output features maps of the two CNNs and feed them to the classifier.
The concatenated features vector is a $4,352$-dimensional vector
obtained by integrating the $4,096$-dimensional output vector of the
\IndNet and the $256$-dimensional output vector of the \PredNet.

In our setting, we train the the \IndNet and \PredNet separately and
exploit the weights of the pretrained convolutional blocks of these
networks to train the fully connected classifier. As for the \IndNet and
\PredNet, we train the model according to a cross-entropy loss function.

\section{Experimental Evaluation}
\label{sec:experimental-evaluation}
This section describes the experimental setup and the tests that have
been carried out to evaluate the robustness of the proposed approach. We
begin describing the dataset and configurations used for this work,
then, in sections \ref{sec:experiments} and \ref{sec:experiments2}
we discuss the results that we obtained on several tests. 

All the experiments discussed in this section were conducted on a Google
Cloud Platform n1-standard-8 instance with 8 vCPUs, 30GB of memory, and
an NVIDIA Tesla K80 GPU. The networks have been implemented using
Pytorch v.1.6 \cite{pytorch}. We trained all the
networks with the learning rate set to $1e-4$, weight decay of the
L2-regularizer set to $5e-5$, and Adam optimizer with an adaptive
learning rate. In our experiments we trained the networks for 80 epochs
with batches of size 32 and early stopping set to 10.

\label{sec:dataset}
To train our model and evaluate its performance, we relied on the VISION
dataset \cite{shullaniVVIDSI2017}. The dataset comprises of 34,427 images
and 1,914 videos, both in the native format and in their social media
version (i.e., Facebook, YouTube, and WhatsApp), captured by 35 portable
devices of 11 major brands.
The dataset has been collected recording 648 native single-compressed
(SC) videos, mainly registered in landscape mode with 
\emph{mov} format. For each device, the videos depict 
flat, indoor, and outdoor scenarios and different acquisition modes. 
The resolution varies from $640 \times 480$ up to $1920 \times 1080$
depending on the device. Furthermore, the dataset contains 622 videos
that were uploaded on YouTube (YT), and 644 shared
through WhatsApp (WA). Similarly to videos, the dataset also contains
images captured in multiple orientations and scenarios and shared via
Facebook 
and WhatsApp. 

In our experiments, we previously process the dataset with the
\emph{ffprobe} \cite{ffmpeg} analyzer from
the \emph{FFmpeg} software to extract the I-frames and P-frames from a
subset of 20 devices. Next, we crop each frame into nonoverlapping
patches of size $H \times W$, where $H = W = 256$, obtaining 153,843
I-frame patches and 209,916 P-frame patches. Finally, we balance all
classes and split the dataset for training, validation, and test with a
proportion of 70\%, 15\%, and 15\%, respectively.

\subsection{Results on Shared Videos}
\label{sec:experiments}

To estimate the performance of our method, we initially
compared the system with respect to a baseline model. Then, we moved
forward to assess the performance of our two-stream architecture, namely
to validate the increase in performance obtained combing the \IndNet and
\PredNet. 

\emph{1) Baseline comparison:} In our first set of experiments we
measured the performance of the single components of \MultiNet
(the \IndNet and \PredNet streams) with
respect to the baseline model introduced by Nam et
al.~\cite{namTSNFDDCHV2019}, for their classification efficacy
when using only I-Frames and P-Frames, respectively. To limit model
training time, we chose to
conduct these experiments on a subset of 10 devices from the VISION
dataset. In fact, in this test, we are not interested in obtaining the
absolute best performances, but we limit ourselves to proving that there
is a boost in performance compared to the baseline. For these
experiments we produce an 80\%-10\%-10\% split of the dataset of the input patches
for training, validation, and test, respectively.

\begin{table}[h!]
\centering
    \begin{tabular}{l c c} 
        \hline
        \textbf{Input} & \textbf{\cite{namTSNFDDCHV2019}} & \textbf{Proposed method}  \\ 
        \hline
        I-Frame& 67.71\% & \textbf{88.42\%} (\IndNet)\\ 
        P-Frame& 67.23\% & \textbf{76.84\%} (\PredNet)\\
        \hline
    \end{tabular}
    \caption{Accuracy on a subset of 10 devices from the VISION
      \cite{shullaniVVIDSI2017} dataset. The proposed method is
	confirmed to be more precise than the baseline at recognizing
	traces left by social networks and apps on frames patches.}
    \label{table:baseline}
\end{table}

The results reported in Table \ref{table:baseline} confirm the
significantly improved
performance of our method respect to the baseline. In fact, the deeper
architectures help to distinguish with higher accuracies (88.42\% and
76.84\% for the \IndNet and \PredNet, respectively) between different
types of double compressions left by social media and messaging apps.
Indeed, the model must be able to distinguish not only between single
and double compression, but also between different types of
double-compression fingerprints. In this sense, a deeper architecture is
capable of extracting more complex information. 

\emph{2) \MultiNet evaluation:} In this test, we evaluate whether and to
what extent our two-stream architecture (\MultiNet) improves even more in terms of
accuracy compared to the single streams.
For this experiment, we trained and evaluated the models on a subset of
20 devices with a dataset split of 70\%, 15\%, and 15\% for training,
validation, and test, respectively. First, we train the \IndNet and
\PredNet in an end-to-end fashion on a subset of 15 devices. Next,
applying transfer learning, we froze the convolutional layers of both
networks and retrained the fully connected classifier on a subset of
5 devices that have not been used on the previous training.
We measure the performance of each network with respect to its accuracy
and its area under the curve (AUC) score.
Table~\ref{table:networks-comparison} reports
the results of these experiments and Table~\ref{table:confusion-matrix}
represents the confusion matrix of
\MultiNet. The experiment confirms that by combining the classification of different
types of frames, the model achieves better performance, with the
\MultiNet gaining up to 95.51\% of accuracy and 96.44\% of AUC
score on patches from SC, WA, and YT. Moreover, the confusion matrix
(see Table~\ref{table:confusion-matrix}) of the \MultiNet on
3,749 patches from 234 unique videos from WA, YT, and SC suggests that
the errors are very small and slightly more numerous in the case of
SC patches. 

\begin{table}[h!]
    \centering
    \begin{tabular}{l r r} 
        \hline
        \multicolumn{1}{c}{\textbf{Model}} & \multicolumn{1}{c}{\textbf{Accuracy}} & \multicolumn{1}{c}{\textbf{AUC}}  \\
        \hline
        \IndNet & 92.32\% & 94.24\%\\ 
        \PredNet & 91.87\% & 93.12\%\\  
        \MultiNet & \textbf{95.51\%} & \textbf{96.44\%}\\ 
        \hline
    \end{tabular}
    \caption{Model accuracies and AUCs on a subset of 20 devices from the VISION dataset \cite{shullaniVVIDSI2017}. The \MultiNet shows higher performance with respect to \IndNet and \PredNet.}
    \label{table:networks-comparison}
\end{table}

\begin{table}[h!]
    \centering
    \begin{tabular}{l|l|c|c|c|}
        \cline{3-5}
        \multicolumn{2}{c|}{}&\textbf{YT}&\textbf{WA}&\textbf{SC}\\
        \cline{2-5}
        \multirow{2}{*}{}& \textbf{YT} & \textbf{1238 (96.41\%)} & $20 (1.65\%)$ & $32 (2.55\%)$\\
        \cline{2-5}
        & \textbf{WA} & $31 (2.41\%)$ & \textbf{1161 (95.79\%)} & $49 (3.91\%)$\\
        \cline{2-5}
        \multirow{2}{*}{}& \textbf{SC} & $15 (1.16\%)$ & $31 (2.55\%)$ & \textbf{1172 (93.53\%)}\\
        \cline{2-5}
    \end{tabular}
    \caption{Confusion matrix of the \MultiNet over YT, WA and SC patches from 234 unique videos of the VISION dataset \cite{shullaniVVIDSI2017}.}
    \label{table:confusion-matrix}
\end{table}

\subsection{Results on Shared Images}
\label{sec:experiments2}
In our last experiment, we measure the robustness of the \IndNet with
respect to images. Specifically, we moved from the intuition that
I-frames are independently encoded using a process similar to JPEG
compression, such that it could be possible to detect images as well as videos
coming from the same social media platform. For this reason we test the
\IndNet trained on videos, on native and WhatsApp images available on
the VISION dataset. Unfortunately, the VISION dataset contains images
uploaded only on WA and Facebook. Therefore, we can apply this test only
on WA images. 
We began the experiment by training the \IndNet on native and WA video
patches obtaining 92.74\% of accuracy. Next, by applying transfer
learning, we froze the convolutional blocks of the network to act as
feature extractors and retrained the fully connected classifier on
images from the same classes. With minimal retraining of the
classifier, it achieves 86.83\% of accuracy. This result suggests that a
mixed method to trace both kinds of media is actually possible.
Therefore, we leave this problem for future research and extensive
experiments.

\section{Conclusions}
\label{sec:conclusions}

In this paper, we introduced a CNN architecture to detect videos
downloaded from social media and messaging apps, based on their content. We
evaluated the advantages of using a deep neural network architecture
and inter-modal relationships between features extracted from different
types of frames. We also explored the possibility of applying
multitask learning to quickly adapt the network from videos to images
obtaining promising results. Future work will take into consideration
new datasets together with multimodal media assets as well as
multitask learning and meta-learning.



\bibliographystyle{IEEEbib}
\label{sec:refs}
\bibliography{strings,refs}

\end{document}